\documentclass[letterpaper, 10 pt, conference]{ieeeconf}

\IEEEoverridecommandlockouts                              
\usepackage{bm}
\usepackage{inputenc}
\usepackage{graphicx}
\usepackage{amssymb}
\usepackage{amsmath}
\usepackage{epstopdf}
\usepackage{multicol}
\usepackage{cite}

\graphicspath{{./images/}}

\usepackage{color}

\title{\LARGE \bf
A Data-Driven Approach to Geometric Modeling of Systems with
Low-Bandwidth Actuator Dynamics
}

\author{Siming Deng$^{1,2,*}$, Junning Liu$^{2}$, Bibekananda Datta$^{2}$, Aishwarya Pantula$^{3}$, David H.~Gracias$^{3}$, \\
Thao D.~Nguyen$^{2}$, Brian A.~Bittner$^{4,\dagger}$, Noah J.~Cowan$^{1,2,\dagger,*}$
\thanks{This material is based upon work supported by the National Science Foundation under grant No.\ \#1830893, EFRI C3 SoRo: Programming Thermobiochemomechanical (TBCM) Multiplex Robot Gels}%
\thanks{$^{1}$Laboratory for Computational Sensing and Robotics, Johns Hopkins University,  Baltimore MD 21218 USA.}
\thanks{$^{2}$Department of Mechanical Engineering, Johns Hopkins University, Baltimore MD 21218 USA.}
\thanks{$^{3}$Department of Chemical and Biomolecular Engineering, Johns Hopkins University, Baltimore MD 21218 USA.}
\thanks{$^{4}$Johns Hopkins University Applied Physics Lab, Laurel,
  MD, 97331  USA.}
\thanks{$^{\dagger}$Co-supervised equally.}
\thanks{$^{*}$To whom correspondence should be addressed: {\tt <sdeng10, ncowan>@jhu.edu}}%
}

\begin{document}

\maketitle
\thispagestyle{empty}
\pagestyle{empty}

\begin{abstract}
  It is challenging to perform system identification on soft robots due to their underactuated, high-dimensional dynamics. In this work, we present a data-driven modeling framework, based on geometric mechanics (also known as gauge theory) that can be applied to systems with low-bandwidth control of the system's internal configuration. This method constructs a series of connected models comprising actuator and locomotor dynamics based on data points from stochastically perturbed, repeated behaviors. By deriving these connected models from general formulations of dissipative Lagrangian systems with symmetry, we offer a method that can be applied broadly to robots with first-order, low-pass actuator dynamics, including swelling-driven actuators used in hydrogel crawlers. These models accurately capture the dynamics of the system shape and body movements of a simplified swimming robot model. We further apply our approach to a stimulus-responsive hydrogel simulator that captures the complexity of chemo-mechanical interactions that drive shape changes in biomedically relevant micromachines. Finally, we propose an approach of numerically optimizing control signals by iteratively refining models, which is applied to optimize the input waveform for the hydrogel crawler. This transfer to realistic environments provides promise for applications in locomotor design and biomedical engineering.

\end{abstract}

\section{Introduction}

Many conventional robots rely predominantly on rigid, fully actuated mechanisms. While these robots maintain superior force and precision compared to natural organisms, these rigid machines usually struggle in tasks that involve interacting safely with humans, handling deformable objects, and operating in unstructured environments \cite{Trivedisoft2008}. Designs from nature have inspired the development of compliant mechanisms in robotics, enabling new capabilities \cite{Shepherdmultigait2011, Shepherddigital2021, tolleymichael2014resilient, liu2019kirigami,  aydin2018design, zhao2015scalable, lu2018bioinspired, chen2022kinegami}.  The emergence of such soft components in robotic platforms has provided new avenues for improved adaptability, safety, cost, and energy efficiency. On the other hand, these new mechanisms introduce new challenges in modeling and control. The compliant nature of these soft components greatly increases the internal degrees of freedom as well as the degrees of underactuation, which makes it hard to obtain precise control of each shape element simultaneously. One potential approach is to utilize variations in passive actuation responses to stimuli across different parts of the body as a means of locomotion. This strategy capitalizes on the distinct temporal dynamics among subsystems or constituents.

In addition to compliance, underactuation, and low-bandwidth dynamics, there are other challenges introduced by soft robotic systems that complicate modeling efforts.  For example, boundary conditions, surface interactions, and nonlinear material properties make it difficult to derive parsimonious models from physical principles. Using a top-down, data-driven approach, this work investigates the aforementioned passive responses within a systematic framework for locomotion control.  We focus on systems with low-bandwidth shape changes in response to a single actuator input. Modeling actuator dynamics and its effects on the system can streamline engineering efforts to design and control soft robots, maximizing their capabilities with less exploratory or exhaustive experimentation. In many cases, a global model of system dynamics may be impractical to obtain, especially for custom machines, machines made through imprecise fabrication techniques, or systems deployed in poorly characterized environments (e.g. a patient's body). A key insight of this work is that in these scenarios, we can avoid reliance on a global model or the sample inefficiency of reinforcement learning schemes by systematically achieving function through iterative modeling and refinement of local behaviors. 

\begin{figure}[tb]
  \begin{center}
    \includegraphics[width=\columnwidth]{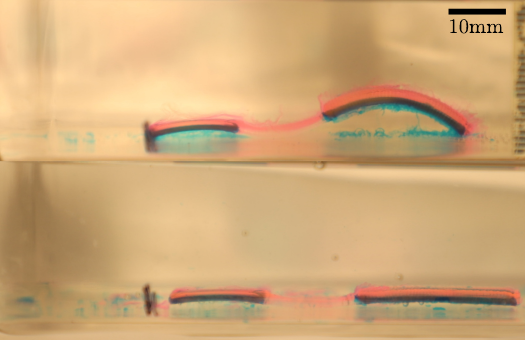}
  \end{center}
  \caption{Experimental screenshots of thermal cycling of two-segment robots with a flexible linker at the end of a cooling half-cycle (top), and heating half-cycle (bottom). One thermal cycle comprises a heating half-cycle and a cooling half-cycle. The robot displaces $4.4\%$ body length at the end of the cycle.}
  \label{fig_crawler}
\end{figure}
%

Geometric mechanics provides a framework through which top-down insights permit novel motion planning approaches to dissipative systems with symmetry, including analytical computation of optimal gaits \cite{Hatton2015, Hattoncartography2017}. A core premise of this work is that complex locomotor mechanics can be rewritten in a kinematic form, owing to the assumption that Rayleigh dissipation dominates the physical interaction between the body and environment \cite{kelly1995geometric}. The same framework has been instrumental in understanding cyclic locomotion in nature \cite{ozkan2017geometric, zhao2022walking}. Bittner et al.\ \cite{Bittner2018} presented a data-driven approach to construct a local model in the neighborhood of the observed limit cycle, using data points from stochastically perturbed, repeated behaviors.  More recent work \cite{Bittnersuds2022} extended this data-driven approach to shape-underactuated systems, which have high bandwidth control available only to a subset of the shape space. The ability to build local models provides the opportunity to sample candidate gaits offline for sample efficient hardware-in-the-loop optimization.

In contrast to hard robots that can readily be powered by multiple, independent electromagnetic modules with rapid (i.e., high-bandwidth) responses, soft robots often must be stimulated by an external signal, typically a single stimulus such as pH, a specific biomolecule, light, or temperature. The impact of this signal must play out through complex kinetics (i.e., \emph{low-bandwidth control}). For this class of systems, shape deformation is not instantaneously coupled to the control signal; rather, there is a temporal lag in the excitation of actuator dynamics following the control input. In our prior work \cite{pantulauntethered2022} (see Fig.~\ref{fig_crawler}). In this prior work, we conceptualized and built a thermo-responsive hydrogel crawler. Although stimulus-responsive shape changes for hydrogels are ubiquitous in literature \cite{breger2015self,han2018soft,na2022hydrogel}, the design of our robot exploited the swelling and shrinking induced bending mechanism, morphological asymmetry, and asymmetry in friction force in response to the change in surrounding temperature to achieve the locomotion. In this crawler, there are three distinct segments: a suspended linker segment connects two end bilayer segments comprising active poly(\textit{N}-isopropylacrylamide) (pNIPAM) and passive polyacrylamide (pAAM) layers with different morphologies. Asymmetry in friction forces, caused by morphological asymmetry, between the two bilayer segments at low and high temperatures allowed the robot to change its anchor during a temperature cycle to move unidirectionally. Additionally, we hypothesize that the distinct swelling rates of these bilayers create such asymmetric ground reaction forces that can be exploited for locomotion. Utilizing the asymmetric response time among segments, this robot can locomote with a single cyclic input---temperature cycle. Alongside the fabrication of this physical crawler, we also developed a Finite Element Analysis (FEA) model \cite{pantulauntethered2022} to simulate the response and investigate the deformation mechanisms, which we use to generate the results in this paper.

Our core contribution, presented in Section~\ref{sec:methods}, is the data-driven modeling of a ubiquitous class of underactuated systems, where the shape dynamics are subject to a band-limited control. This contrasts with prior work \cite{Bittnersuds2022}, which required the application-limiting assumption that at least one element can be accessed by high bandwidth control. Our work enables the application of our framework for efficient behavior optimization and enhancement across a wide spectrum of previously unexplored soft robotic systems, such as hydrogel crawlers \cite{Bittnersuds2022}.
In ~\ref{sec:swimmer}, we demonstrate our methods on a well-known, analytically tractable system, modified to include low-bandwidth actuation of its shape parameters. Finally, in Section~\ref{sec:application} we test our methods on a high-dimensional, finite element model of our previously published hydrogel robot \cite{pantulauntethered2022}. In both examples, we show how the actuator dynamics can be simultaneously modeled with the body movements, enabling a data-driven modeling architecture for a broader class of soft or underactuated systems.  Further, we use these examples to numerically optimize a parameterized input signal for certain objectives using an iterative parameter optimization and model refinement approach.

\section{Background}

\subsection{Geometric locomotion model}
Geometric mechanics \cite{kelly1995geometric,cendra2001geometric, BlochNonholonomic2003} provides a framework for locomotion based on exploiting symmetry. In this framework, a distinction is made between the internal configuration (shape) of a locomotion system and its position and orientation (group) in a spatially fixed reference frame. Central to this framework is the idea of group invariance of the dynamics \cite{ostrowskioptimal1998}: a shape change that moves the system in a certain way (relative to the body frame) will have the same effect, invariant to the absolute position and orientation.

Here, we consider a subclass of such group-invariant dynamical systems that are dominated by Rayleigh dissipation as caused by many types of isotropic friction \cite{kelly1996geometry}; in such dissipation-dominated systems, the equations of motion can be kinematically reduced such that the body velocity is expressed as a shape-dependent linear mapping of shape velocity. In this case, the kinematic equation is
\begin{equation}
  \label{eq_reconstruction}
  \big(g^{-1}\dot{g}\big)^\vee = \xi = -\bm{A}(r)\dot{r},
\end{equation}
where $\xi$ is the group velocity in its body frame\footnotemark[1], $r$ denotes the system shape, and $\bm{A}(r)$ is called the \textit{local connection}. Here the matrix $\bm{A}(r)$ is a function of shape $r$ and acts analogously to a Jacobian in which it relates the system's shape velocities to body velocities. A spatial trajectory of the system body frame can be calculated by integrating \eqref{eq_reconstruction} with respect to a fixed reference frame.

\footnotetext[1]{Here $(\cdot)^\vee$ is an isomorphism that maps
  velocities from Lie algebra form to vector form, and its
  inverse is denoted as $\widehat{(\cdot)}$. In a SE(2),
  $(\cdot)^\vee: \mathrm{se}(2) \rightarrow \mathbb{R}^3$, and
  $\widehat{(\cdot)}: \mathbb{R}^3 \rightarrow \mathrm{se}(2)$}

For systems within the scope of \eqref{eq_reconstruction}, the local connection can be analytically derived from a set of Pfaffian constraints on the system's shape and body velocities. A global model can be empirically estimated by exhaustively sampling the system shape space and its tangent bundle (the collection of shape velocities available at each point in the shape space) \cite{Dai2016GeometricSO}. However, such global models are often difficult to obtain for animals or underactuated systems because of the challenges in sampling this space with sufficient density.

\subsection{Data-driven modeling} \label{datadriven} 
Bittner et al.~\cite{Bittner2018} developed a data-driven approach to geometric modeling and optimization, which was later extended \cite{Bittnersuds2022} to be applied to systems with high bandwidth control in only a subset of the shape variables. This approach allows local estimation of a connection in the neighborhood of a limit cycle with far fewer samples than required to train a global model, making it practical for \textit{in-situ} system identification, especially for systems with high dimensional shape spaces.

In this approach, system shape data (in the form of a regularly sampled time series) are fit to an oscillator such that each data point is assigned a phase value \cite{Revzenestimate2008}. A zero-phase-lag Butterworth smoothing filter is applied before finite differencing to obtain time derivatives of both shape and position. Then, a local Taylor approximation of the connection can be computed via linear regression across data points within phase windows. A Fourier series is then fit to these local regression coefficients to build a model that is supported at any queried phase.

We detail the process by which we estimate a linearized model within each phase window.  Data-driven Floquet analysis techniques extract information from the observed oscillator data and assign each sample point an estimated phase \cite{Revzendatadriven2015, Revzenestimate2008}.  The observed shape samples are then phase-averaged and fitted to a Fourier series to obtain a limit cycle, denoted as $\theta_r(\cdot)$. The perturbed trajectory, relative to the limit cycle, is denoted as $\delta_r:= r - \theta_r$. The first-order Taylor approximation of the local connection in each local phase window can be constructed as
\begin{equation}
  \label{eq_bvmodel}
  \bm{A}_k(r) \approx \bm{A}_k(\theta_r) + \delta_r^T\frac{\partial \bm{A}_k}{\partial r},
\end{equation}
where $\bm{A}_k(r)$ is the $k^\mathrm{th}$ row of the local connection, which is a vector of the same dimension as shape perturbation $\delta_r$. 

All samples are grouped into neighborhoods by their estimated phase values, and a local model is fitted in each phase window. In the $m^\mathrm{th}$ phase window, the averaged shape is assumed to be a constant $\theta_r^m$. The first-order Taylor approximation of the local connection matrix $\bm{A}(r)$ in this phase window can be fitted by solving the following Generalized Linear Model (GLM):
\begin{equation}
  \label{eq_glm}
  \xi^{(n)}_k \sim \bm{C}_k + \bm{B}_k\delta^{(n)}_r + \bm{A}_k(\theta_r)\dot{\delta}^{(n)}_r + \frac{\partial \bm{A}_k}{\partial r}\delta_r^{(n)}\dot{\delta}^{(n)}_r.
\end{equation}
Here, $\xi^{(n)}_k$ corresponds to the $k^\mathrm{th}$ coordinate of the $n^\mathrm{th}$ sampled body velocity $\xi^{(n)}$, and $\delta^{(n)}_r:=r^{(n)}-\theta_r^m, \dot{\delta}^{(n)}_r:=\dot{r}^{(n)}-\dot{\theta}_r^m$ are the shape and shape velocity perturbation samples defined in the local region indexed by $m$.  Regressor $C_k := \bm{A}_k(\theta_r)\dot{\theta_r}$ describes the average behavior in the neighborhood of $\theta_r^m$. $B_k:= \dot{\theta}_r^T\frac{\partial \bm{A}_k}{\partial r}$ and $\bm{A}_k$ are the terms that respectively relate the effects of shape and shape velocity offsets from the limit cycle. $\frac{\partial \bm{A}_k}{\partial r}$ is the cross term that incorporates the interaction between $\delta_r$ and $\dot{\delta}_r$. Note that this is a local estimate in the $m^\mathrm{th}$ phase window. This local approximation is repeated for all separate groups of data points, after which a Fourier series model is used to guarantee a smooth transition among the fitted matrices.


\section{Methods} \label{sec:methods}

\subsection{Low-bandwidth shape control}

In this paper, we consider systems whose locomotion can be characterized by \eqref{eq_reconstruction} while only having access to low-bandwidth control over $r$. In particular, we assumed the dynamics on $r$ to take the general form of
\begin{equation}
  \label{eq_actuation}
  \dot{r} = f(r,u),
\end{equation}
where the system shape velocity $\dot{r}$ is a function of its shape $r$ and an input $u$.

First, we extracted a phase-averaged gait cycle $(\theta_r,\theta_u)$ from the general input \cite{Revzenestimate2008}. Denoting the perturbation from phase-averaged shape and control as $\delta_r:= r - \theta_r, \delta_u:= u- \theta_u$, the local first-order Taylor approximation of the actuation dynamics can be written in the following form: 
\begin{equation}
  \label{eq_actuatorregressor}
  f(r,u) \approx f(\theta_r,\theta_u)+\frac{\partial f}{\partial r}(\theta_r,\theta_u)\delta_r+\frac{\partial f}{\partial u}(\theta_r,\theta_u)\delta_u
\end{equation}
We then fit the data to the above first-order approximation by solving the following Generalized Linear Model,
\begin{equation}
  \label{eq_actuatorregressor_letter}
  \dot{\delta}^{(n)}_r \sim \bm{D} + \bm{E}_r \delta^{(n)}_r + \bm{E}_u \delta^{(n)}_u,
\end{equation}
where $D$ is the average shape velocities of the observed data in the local phase window, and ($\bm{E}_r, \bm{E}_u$) are the terms that describe how shape and input offsets respectively modify the average behavior. $\delta^{(n)}_r := r^{(n)} - \theta^m_r, \delta^{(n)}_u:= u^{(n)}- \theta^m_u$ are the shape and input perturbations defined in the $m^{\mathrm{th}}$ local phase window, where $(\theta^m_r,\theta^m_u)$ are the mean values of shape and input. 

The estimation of the local connection can be done separately from the actuator dynamics, hence this part remains identical as in \ref{datadriven}. We repeat the same procedure for all discrete phase windows and use a Fourier series to smoothly connect all local models.

The fitted models from \eqref{eq_actuatorregressor_letter} and \eqref{eq_glm} can be used in series to make predictions of the system shape and position trajectories given the input signal. First, the input signal $u(t)$ is transformed into the phase coordinate using the fitted phase map. The initial shape is assumed to be on the limit cycle ($\delta_r=0$) as the same phase value of the initial input $u(t_0)$. At each discrete time $t_i$, the shape velocity perturbation $\dot{\delta}_r(t_i)$ is predicted using the actuator model \eqref{eq_actuatorregressor_letter} given the current shape perturbation $\delta_r(t_i)$ and the input perturbation $\delta_u(t_i)$. $\delta_r(t_i)$ is then integrated by Euler's method to obtain the predicted shape at the next time step $\delta_r(t_{i+1})$. The predicted shape $\delta_r(t_{i})$ is then used to predict the body velocity $\xi(t_{i})$ using the body velocity model \eqref{eq_glm}. The predicted body frame position at the next time step $g(t_{i+1})$ is then integrated using $\xi(t_i)$.

When building an actuator model, the system shape trajectory is the integrated estimation of the shape velocity predictions. Simultaneously, it also appears as the input to the local connection in the locomotion model. As in prior work, the shape and body motion models are predicted in separate stages. Note that in the process of simulating a system spatial trajectory from a general input signal, the two integration steps of each model evolve in series. We start with knowledge of the initial system shape $r(t=0)$ and the control input $u(t)$. Then we can numerically solve \eqref{eq_actuation} and \eqref{eq_reconstruction} together using the fitted regression models, \eqref{eq_actuatorregressor_letter} and \eqref{eq_glm}.

We apply the model improvement metric described in \cite{Bittnersuds2022}, comparing our first-order regression model predictions to the phase-averaged baseline model predictions,
\begin{equation}
  \label{eq_metric}
  \Gamma_{\chi} = 1-\frac{\sum^\mathcal{N}_{n=1}\|\chi_D^{(n)}-\chi^{(n)}\|}{\sum^\mathcal{N}_{n=1}\|\chi_T^{(n)}-\chi^{(n)}\|}.
\end{equation}
This improvement metric is defined as one minus the relative error of the data-driven prediction $\chi_D$ with respect to the baseline prediction $\chi_T$ over $\mathcal{N}$ samples of body velocity and shape velocity $\chi = \{\xi,\dot{r}\}$. $\Gamma_{\chi} \leq 0$ means the data-driven prediction is no better than the phase-averaged prediction, and $0 < \Gamma_{\chi} \leq 1$ means that our model can make better predictions than the baseline model, up to perfect reconstruction of the ground truth at $\Gamma_{\chi} = 1$.

The baseline phase model corresponds to the zeroth-order model, generating predictions solely reliant on current phase information. This metric holds significance in assessing the extent to which our data-driven first-order model outperforms the baseline model within the local region. A substantial improvement metric implies that the data-driven model more accurately approximates the ground truth compared to the baseline model within the local region, making it more suitable for local optimization.

\subsection{Optimizing behaviors and iterative model refinement}
Once an initial model is obtained, we can make predictions on the system position trajectories $g(t)$ by a general control input $u(t)$. Using finite differencing, we can estimate the gradient of displacement per cycle with respect to the control parameters around the observed data. We can then utilize the estimated gradient and Hessian to numerically optimize the control parameters for certain behaviors of the system (e.g. maximizing the displacement per cycle).

The expense of data collection\footnotemark[1] incentivizes our focus on sample-efficient optimization schemes. Given an input parameterization, we sparsely sample data in the full range of the input space and build a rough model. According to this rough model, we numerically optimized the input parameters for certain behavioral objectives. Then we zoom into the region around the optimized parameter and re-sample points in this local area. The model built with sample points in a smaller region will be more localized and accurate over that confined domain. We iterate between these two processes---optimization and model refinement---so that in the end it converges to a local optimum in the control space. A global optimum is not guaranteed. In Sections \ref{sec:swimmer} and \ref{sec:application}, we demonstrate the methods using a sample objective where we maximize the displacement per cycle and penalize the cycle time.

\footnotetext[1]{In practice, a typical temperature cycle for our hydrogel crawler takes approximately 6 hours because of the slow actuation kinetics of the material. The FEA simulation is computationally expensive as well; running a 10-cycle simulation on a well-equipped desktop computer takes about 2 days. Both facts make data collection for such systems expensive, thus data efficiency is crucial.}

\section{Illustrative example: Purcell swimmer with low-bandwidth
  actuation} \label{sec:swimmer} 
\begin{figure}[ht]
  \begin{center}
    \includegraphics[width=\columnwidth]{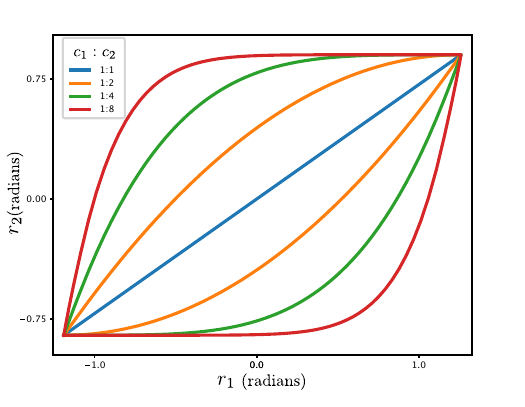}
  \end{center}
  \caption{Plotted on the shape space, phase-lag-induced shape trajectories exhibit the hysteresis often observed in low-bandwidth control systems. The greater the difference in the reaching rate $c_i$, the larger the phase lag is. This behavior emerges by waiting for a sufficient amount of time for the two joints to both reach steady-state equilibrium and then exciting both joints by a step change in input.}
  \label{fig_lpphase}
\end{figure}

Before demonstrating the method on data, it is helpful to interrogate a simple analytical model, such as a three-link Purcell swimmer.  We modify the model to include low-bandwidth actuation, inspired by the low-bandwidth actuation of our hydrogel robot:
\begin{equation}\label{eq_dynamics}
  \dot{r_i} = c_i(r_{i}^s(T)-r_i),\quad c_i>0,i=1,2,
\end{equation}
where $r_i$ is the $i^\mathrm{th}$ shape variable (joint positions), $r_{i}^s (T)$ denotes each joint's steady state equilibrium given temperature, and $c_i$ is the converging rate of each joint towards its steady state equilibrium. Specifically, the steady-state equilibrium $r_{i}^s (T)$ is assigned to be a linear function of a one-dimensional input signal, temperature $T$. We have a bound on temperature that puts limits on the swimmer's joint angles. The resulting shape trajectory exhibits hysteresis which is often observed in low-bandwidth control systems (see Fig.~\ref{fig_lpphase}).

Assuming different constants $c_i$ on the two joints, the shape variables will exhibit a gait where both joints are not synchronized under a repeating temperature cycle, see Fig.~\ref{fig_input}. Although both joints are controlled by the same temperature input, the phase lag between the two joints breaks the symmetry of joint synchronization, making the gait enclose a nonzero area in the shape space, which is critical for locomotion in viscous swimming domains from the scallop theorem \cite{purcell1977life}.
\subsection{Input generation} \label{toy_input} 
\begin{figure}[tb]
  \begin{center}
    \includegraphics[width=\columnwidth]{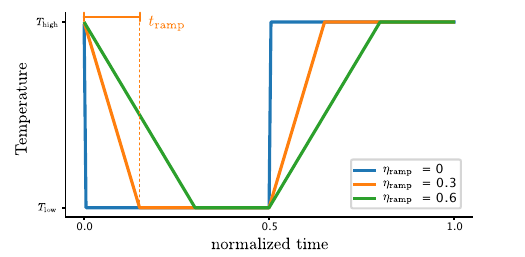}
  \end{center}
  \caption{Example temperature cycles (our low-bandwidth control input), normalized for period and temperature, where $t_{\mathrm{ramp}} = \eta_\mathrm{ramp}\cdot t_\mathrm{cycle}/2$ is the time used to ramp input to the goal temperature. Since the ramp time should always be smaller than half of the period, the use of the ramp time ratio avoids additional constraints other than a bounding box. When $\eta_{\mathrm{ramp}}$ is small, the temperature profile is more square-wave-like, and the resulting shape trajectory is more likely to have a larger phase lag. Consequently, for longer ramp times, the shape trajectory is more likely to have minimal phase lag because both joints are in equilibrium for the duration of temperature change.}
  \label{fig_input}
\end{figure}

\begin{figure}[tb]
  \begin{center}
    \includegraphics[width=\columnwidth]{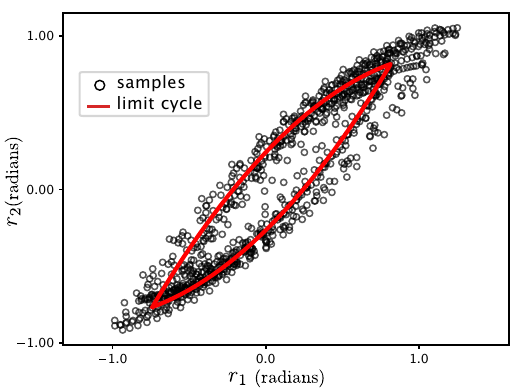}
  \end{center}
  \caption{Sampled shape data points around a limit cycle. Note that we cannot directly sample the full shape and shape velocity space due to the presence of the actuator dynamics. Fortunately, adding perturbation to the input parameters does generate shape and shape velocity variations around the limit cycle, which is essential for building models of behaviors.}
  \label{fig_orbit_var}
\end{figure}

Our parameterization on the control signal is concise while maintaining the ability to alter important features of the temperature profile.  Here, we used 4 parameters to describe the temperature cycle: a low-point temperature $T_\mathrm{low}$, a high-point temperature $T_\mathrm{high}$, time-span per cycle $t_\mathrm{cycle}$, and the portion of the half period to ramp between high and low temperatures $\eta_{\mathrm{ramp}} = 2t_\mathrm{ramp} / t_\mathrm{cycle}$, where $t_\mathrm{ramp}$ is the time to ramp between high and low temperatures, see Fig.~\ref{fig_input}.
Performing multiple cycles of these parameterized temperature cycles, the shape trajectory forms a stable orbit under periodic forcing. We then perturbed the forcing parameters across cycles, which resulted in what can be seen as a "tube" around the orbit as shown in Fig.~\ref{fig_orbit_var}.
\section{Main application: hydrogel crawler} \label{sec:application}

\subsection{Actuator dynamics}

Bilayers and other multi-material structures are useful in creating interesting modes of shape changes like bending. Typical swelling-driven bilayer bending dynamics are similar to the form of an exponential low-pass filter as shown in Fig.~\ref{fig_lpphase}. Specifically, the geometry of the bilayer (e.g., layer thickness ratio and material properties) can affect the steady state equilibrium of the shape variables as well as the rate of reaching equilibrium. 
%
%

\subsection{Hydrogel crawler}
Thermo-responsive hydrogel crawlers in \cite{pantulauntethered2022}, capable of swelling and shrinking, utilize geometric asymmetry, leading to asymmetry in friction forces, to generate net motion under temperature cycles. We utilized the same 2D finite element model in Abaqus Unified FEA \cite{AbaqusFEA} to produce a time-dependent $x-y$ coordinate along the contour of the robot to calculate the area (2-D volume) of the actuated segments. The data is then parameterized into the shape variable $r$.

Here we assume the actuation dynamics in a general (nonlinear) form 
\eqref{eq_actuation}, without any specific structure on it, namely
\begin{equation}
  \label{eq_generaldyn}
  \dot{r} = f(r,T),
\end{equation}
where the input is assumed to be the temperature $T$.




\subsection{Finite element model}

Briefly, our finite element model, based on chemo-mechanics described in \cite{chester2011gel}, solves coupled diffusion-deformation equations for hydrogel undergoing temperature-driven swelling and shrinking. We used Neo-Hookean and Flory-Huggins potentials to describe the entropic elastic behavior of the polymer network and the mixing of polymer-solvent, respectively. The swelling of pNIPAM caused by the lower critical solution temperature transition (LCST) was modeled by assuming a sigmoidal function for the temperature dependence of the Flory-Huggins interaction parameter. We also assumed that the diffusivity of the water through pNIPAM also increased sigmoidally with temperature across the LCST, which caused the characteristic time of deswelling to be significantly faster than the characteristic time of swelling. Given the relatively long actuation time, the environment temperature can be simply assumed to be evenly distributed.  We also considered a combined effect of gravity and buoyancy by prescribing a net body force on the hydrogel structure. Our material model included a total of 10 parameters which were either directly determined from experiments or calibrated against experiments using finite element analysis. A list of the parameters and their values are listed in Table \ref{tab_mat_props}. In addition, we assumed a rigid frictional surface with a friction coefficient, $\mu_k = 0.1$, underneath the robot to facilitate friction-driven locomotion induced by geometric asymmetry. Further details of the finite element simulation can be found in \cite{pantulauntethered2022}.

\begin{figure}[tb]
  \begin{center}
      \includegraphics[width=\columnwidth,trim={0cm 5cm 0cm 5cm},clip] {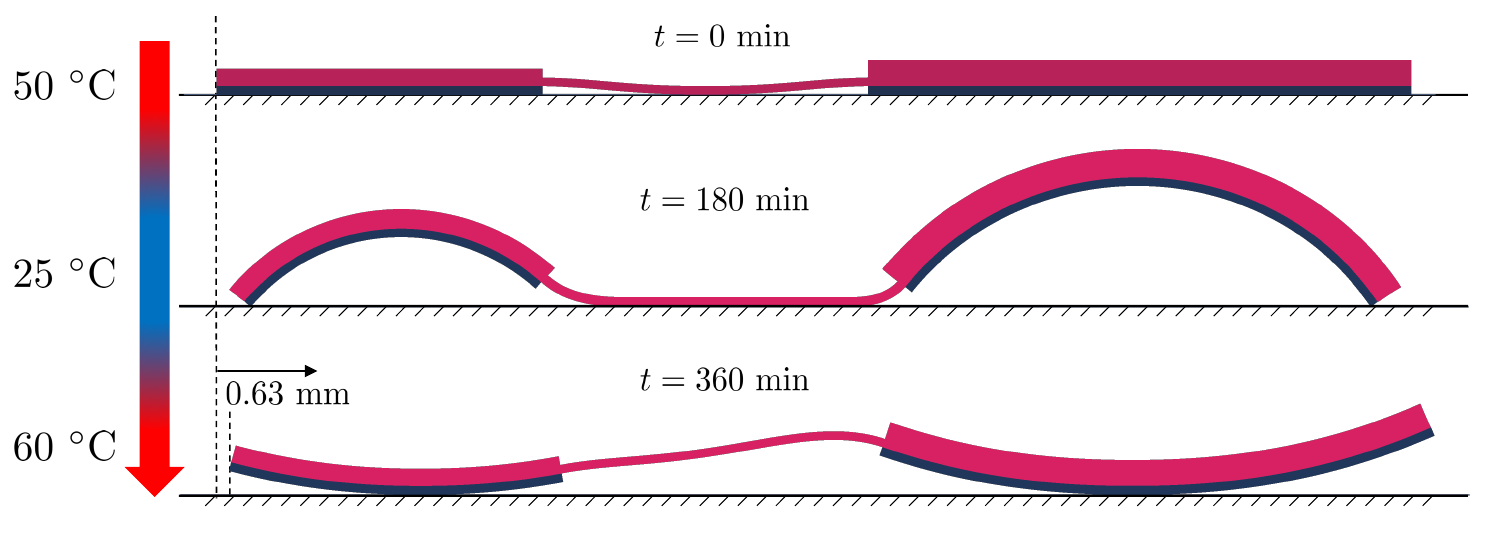}
      \caption{A representative finite element analysis showing unidirectional motion towards the larger bilayer of the thermo-responsive hydrogel crawler subjected to 3 hours of ramped cooling and 3 hours of ramped heating cycle.}
      \label{fig:fea_robot}
  \end{center}
\end{figure}

\begin{table}[tb]
  \begin{center} 
    \begin{tabular}{||c|c||}
      \hline
      Parameter & Value \\
      \hline
      Differential density between polymer and water, $\Delta \rho$ & 100 kg/m$^3$\\
      Shear modulus of pNIAPM, $G_{\mathrm{pNIPAM}}$ & 2.32 kPa\\
      Shear moduli ratio of the gel, $\frac{G_{\mathrm{pAAM}}}{G_{\mathrm{pNIPAM}}}$ & 15.36 \\
      Curing temperature, $T_{\mathrm{cure}}$ & 323 K \\
      Flory-Huggins parameter at high temperature, $\chi_H$  & 0.8031 \\
      Flory-Huggins parameter at low temperature, $\chi_L$ & 1.0483 \\
      Transition temperature, $T_{\mathrm{trans}}$ & 318 K \\
      Range of transition temperature, $\Delta$ & 8.0\\
      Diffusivity at low temperature, $D_{\mathrm{L}}$ & 10$^{-11}$ m$^2$/s \\
      Diffusivity at high temperature, $D_{\mathrm{H}}$ & 10$^{-10}$ m$^2$/s \\
      \hline
    \end{tabular}
    \caption{List of materials properties used in the simulation}
    \label{tab_mat_props}
  \end{center}
  \vspace{-2em}
\end{table}

\subsection{Input generation}
Here we used a similar parameterization for the input temperature signal as that in \ref{toy_input}. However, for thermo-responsive hydrogels, the kinetics of the swelling and shrinking vary distinctively. We, therefore, separated the input cycle into two independent parts, cooling and heating. Thus, the dimension of the input parameter space increases to six, low temperature $T_\mathrm{low}$, ramped cooling time span $t_\mathrm{cool}$, cooling ramp time ratio $\eta_\mathrm{cool}$, high temperature $T_\mathrm{high}$, ramped heating time span $t_\mathrm{heat}$, and heating ramp time ratio $\eta_\mathrm{heat}$. The ramp time ratios are defined as the ratio of the corresponding ramp time to the cooling or heating time span, i.e., $\eta_\mathrm{cool} = t_\mathrm{\{ramp, cool\}}/t_\mathrm{cool}$. The allowable range of each parameter is determined by material properties and characteristic diffusion time and was validated by a parametric study using FEA. Specifically, the ramped cooling and heating timespans are determined by scaling swelling and shrinking characteristic time of the hydrogel, both temperature ranges are specified by $4\%$ equilibrium strain span of the material. The ramp ratios are ideally in the range of $[0,1]$, but small ramp ratios mean very large rates of temperature change, which is impractical and often causes numerical stability and convergence issues in FEA simulation because of the excessive deformation of the finite element mesh in a short period of time. Thus in the implementations, we raised the lower bounds to $\frac{1}{32}$. The calculated full input parameter ranges are shown in Table \ref{tab_input_range}. The noisy input signal is generated by sampling from a uniform distribution in the parameter space. The input parameters are then used to generate noisy temperature cycles for FEA. To avoid numerical issues, instead of running hundreds of thermal cycles at a time, each FEA simulation comprised 10 thermal cycles. At each iteration, we ran 10 simulations resulting in 100 cycles of input data for our data-driven model.

\begin{table}[tb]
  \begin{center} 
    \begin{tabular}{||c|c||}
      \hline
      Parameter & Range \\
      \hline
      $T_\mathrm{low}$ & $20\sim41$  $^\circ C$ \\
      $T_\mathrm{high}$ & $45\sim65$  $^\circ C$ \\
      $t_\mathrm{cool}$ & $2\sim8$  hrs \\
      $t_\mathrm{heat}$ & $0.5\sim3$  hrs \\
      $\eta_\mathrm{cool}$ & $\frac{1}{32}\sim1$ \\
      $\eta_\mathrm{heat}$ & $\frac{1}{32}\sim1$ \\
      \hline
    \end{tabular}
    \caption{Full input parameter ranges.}
    \label{tab_input_range}
  \end{center}
\end{table}

\begin{figure*}[tb]
  \begin{center}
    \includegraphics[width=\textwidth]{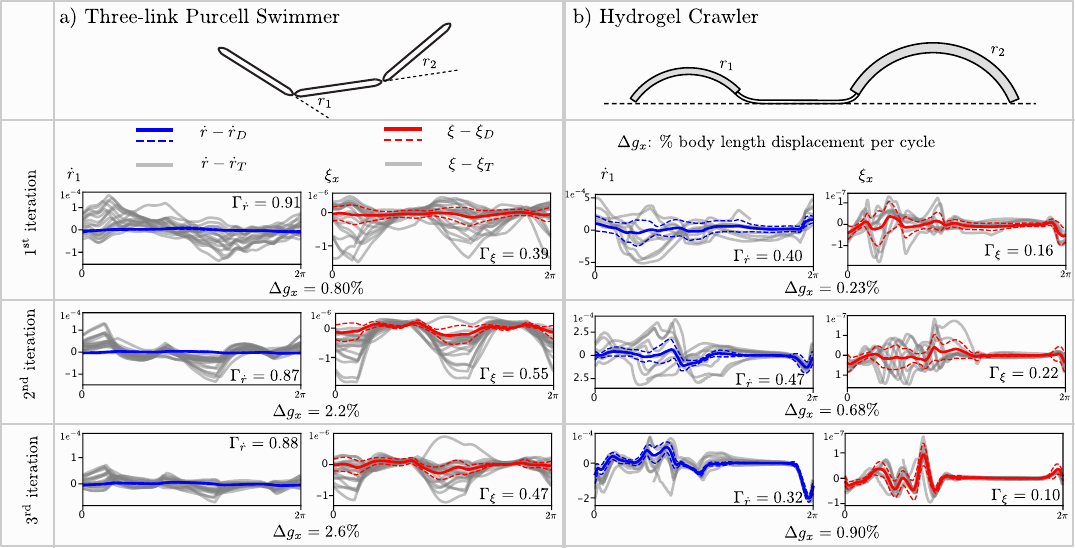}
  \end{center}
  \caption{Model prediction metrics for the Purcell swimmer (top left) and hydrogel crawler (top right) model. We plot sample model prediction errors in phase coordinates from each of the model refining iterations. Also, the average displacement, in terms of percentage body length per actuation cycle, each iteration is shown below the plots. The metrics are calculated on the testing set which is unseen by the model both on shape velocity predictions and body velocity predictions. Test set trajectories (grey) along with the mean (solid) and standard deviation (dashed) of the data-driven model prediction error are plotted for each iteration. The body velocity prediction error is inherently propagated from the shape velocity prediction error, and thus the body velocity prediction errors are unsurprisingly greater than the shape velocity prediction error. While the model prediction errors start to drop as we shrink the sampling range due to overfitting, the average behavioral objective improves each iteration.}
  
  \label{fig_metrics}
\end{figure*}

\subsection{Shape parameterization}
The compliant nature of the devices and external forces makes the internal shape high-dimensional. However, fitting models to a highly dimensional shape space will likely cause overfitting. We thus seek a reduced-order representation of the shape of the system.  Here, principal components analysis (PCA) is a simple candidate reduction method that could serve this purpose. We tried PCA on the streamline along the crawler body, from which we calculated the internal angles between each of the segments. Fitting a PCA model on the internal angles, we found that the first two principal components (modes) reconstruct over 90\% of the data. While it is a straightforward way of reducing the effective degrees of freedom, the complex coupling between segments led to principal components that lacked a clear physical interpretation. As an alternative, we considered that the volume of each active section is a more physical, descriptive candidate for representing the system shape variables. To do this, we estimated the volume at each time point based on contour points from the FEA simulation, and it was used to calculate the enclosed area (2D volume for our planar FEA). This parameterization provided a clear, intuitive relationship between the two segments, and exhibited the phase lag between the smaller and larger bilayer segments that we expected from the design.

\subsection{Input optimization}
As a demonstration, we optimized the input parameters to maximize the
displacement per cycle. The objective function is defined as
\begin{equation}
  \label{eq_objective}
  F(u) = \Delta g_x - \lambda t_{\mathrm{cycle}},
\end{equation}
where $\Delta g_x$ is the displacement in the $x$ direction per cycle, and $t_{\mathrm{cycle}}$ is the cycle time. $\lambda$ is a penalty factor that controls the trade-off between the above two terms. The objective function is to maximize the net displacement per cycle. During the optimization process, we noticed that the optimizer tended to find cycles with the longest possible cycle times, optimizing cycle-to-cycle distance, with no penalty on time, pushing the results toward the boundary. To address this, we added a regularizing term to penalize the cycle time.

We started by sampling 100 points (resulting in 100 cycles of system motion) in the full input parameter space as shown in Table \ref{tab_input_range}. We performed ten-fold cross-validation to avoid overfitting. A rough model of the system was built using the initially sampled data points, and then the model was used to optimize for an input parameter that maximizes the objective function above. The optimization was performed using the Sequential Least Squares Programming algorithm where the local gradient and Hessian were estimated using finite differencing. Once the numerical optimum was obtained, we shrank the range of each input parameter by $35\%$, centering at the optimum, and repeated the optimization process. We repeated this process three times, and the parameters converged to a performant gait. The model improvement metrics were calculated for each iteration as shown in Fig.~\ref{fig_metrics}

\section{Discussion and Conclusion}

In this work, we designed and implemented a data-driven modeling framework for dissipative systems with low-bandwidth actuator dynamics. We showed the success of this method in predicting behaviors on a classical simplified model, the Purcell swimmer, with a modified class of passive shape dynamics. We built on prior work, relaxing the requirement that at least one shape element is accessed through high-bandwidth control. In doing so, our method enables the modeling of novel mechanisms like the hydrogel crawler, whose internal degrees of freedom all exhibit a passive response to controllable stimuli. Despite the challenges often associated with designing control signals under low-bandwidth actuation, the robot was intentionally designed to capitalize on morphology-induced actuation asymmetries for locomotion.  We showed not only that we could model the crawler with accuracy beyond the phase-averaged gait, but that the system was capable of using this model in a gradient-based optimization scheme to rapidly identify a viable crawling maneuver. The broader implications of this result are that we now have a rational framework to pursue data-driven modeling and optimization of a much larger class of underactuated systems. For applications in biology, where continuous, soft interfaces facilitate safe interaction with the body, this method provides the potential to model new mechanisms pre-deployment in the body and even \textit{in situ}, since variation amongst morphology and environment across patients can be significant. Key additional future efforts will include power, actuation, and sensing at the scales desired for the locomotion application. In addition, it is well known that there is a significant phase lag between muscle activation and body movements \cite{Goldmanemergence2013}, suggesting that our approach can be used to better understand the underlying neural control problem~\cite{madhavsynergy2020}.


The model improvement saturated and decayed as the sampling region was reduced in the hydrogel crawler gait optimization result. This likely occurred because there was less variation in the sampled data (our tube of gait distribution data had a smaller overall volume). This is loosely analogous to convergence results in adaptive control, which often rely on sufficient excitation of the dynamics; in a similar vein, we do not expect to learn informative improvement without cycles that excite significant dynamic variation.

We have shown that the model was iteratively improved and the optimizer settles at a performant gait. In principle, brute force sampling of initial conditions could be used to identify a variety of locally optimal gaits from which the best one could be selected. In this project, a single cycle of FEA simulation takes hours to run, making it impractical to exhaustively sample all initial conditions in this way. Likewise, reduced order modeling is challenging for many next-generation soft robots a priori, (e.g., custom morphologies deployed in uncharacterized patient bodies such as an artificial heart valve \cite{Hasan2016heartvalve}). Thus, it is crucial to be able to build a viable control policy from limited data, which was our primary objective for the hydrogel crawler. We have provided a framework through which it and similar robots can rapidly obtain functional motion primitives. We juxtapose our method with reinforcement learning \cite{Erden2008freegait}, deep learning architectures \cite{Tsounis2020deepgait}, and gait optimization \cite{winkler2018gaitopt} which require large amounts of data that are unavailable in simulation and impractical to empirically obtain for many applications (e.g.\ in biomedical settings).

While many samples might be available in simulated environments, there are many platforms that must be system-identified in the field. \textit{In-situ} system identification (such as the type we implement here) paired with a gradient-based optimizer provides a tractable, system-oriented way to pursue optimization of robot behaviors in hard-to-model environments. Further efforts can be made to wrap control, modeling and optimization into an in-the-loop process, where the robot can learn to navigate in complex, time-varying environments without human intervention.


\bibliography{IEEEabrv,reference}

\end{document}